%% file: TTPA.tex
\pdfoutput=1
\documentclass[11pt]{article}
\usepackage{arydshln}
\usepackage{adjustbox}
\usepackage[dvipsnames]{colortbl}
\definecolor{Gray}{gray}{0.95}
\usepackage[inline]{enumitem}
\usepackage{enumerate}  
\usepackage{hyperref}
\usepackage{float}
\usepackage[preprint]{acl}
\usepackage{multirow}
\usepackage{times}
\usepackage{latexsym}
\usepackage{booktabs} 
\usepackage[T1]{fontenc}
\usepackage{array}
\usepackage[utf8]{inputenc}
\usepackage{amsmath}
\usepackage{microtype}
\usepackage{mathrsfs}
\usepackage{amsthm,amssymb}
\usepackage{inconsolata}
\usepackage{xspace}
\usepackage[many]{tcolorbox}
\usepackage{graphicx}

\newcommand{\eg}{\emph{e.g.,}\xspace}

\newcommand{\ignore}[1]{}

\newcommand{\fullmodel}{\textbf{T}oken-level \textbf{T}ool-use \textbf{P}reference \textbf{A}lignment Training Framework\xspace}
\newcommand{\model}{TTPA\xspace}
\newcommand{\reward}{ESM\xspace}
\newcommand{\sample}{TPS\xspace}

\title{TTPA: Token-level Tool-use Preference Alignment Training Framework with Fine-grained Evaluation}

\author{
Chengrui Huang$^{1}$, Shen Gao$^{1}$, Zhengliang Shi$^{2}$, \\
{\bf Dongsheng Wang$^{1}$, Shuo Shang$^{1}$\thanks{Corresponding author.}} \\
$^{1}$ University of Electronic Science and Technology of China,\\
$^{2}$ Shandong University\\
\texttt{ry.cr.huang@gmail.com},
\\ 
\texttt{shengao@uestc.edu.cn},
\texttt{jedi.shang@gmail.com}
}

\begin{document}
\maketitle
\begin{abstract}
Existing tool-learning methods usually rely on supervised fine-tuning, they often overlook fine-grained optimization of internal tool call details, leading to limitations in preference alignment and error discrimination. 
To overcome these challenges, we propose \fullmodel (\model), a training paradigm for constructing token-level tool-use preference datasets that align LLMs with fine-grained preferences using a novel 
error-oriented scoring mechanism. 
\model first introduces reversed dataset construction, a method for creating high-quality, multi-turn tool-use datasets by reversing the generation flow. 
Additionally, we propose \textbf{T}oken-level \textbf{P}reference \textbf{S}ampling (\sample) to capture fine-grained preferences by modeling token-level differences during generation. 
To address biases in scoring, we introduce the \textbf{E}rror-oriented \textbf{S}coring \textbf{M}echanism (\reward), which quantifies tool-call errors and can be used as a training signal. 
Extensive experiments on three diverse benchmark datasets demonstrate that \model significantly improves tool-using performance while showing strong generalization ability across models and datasets. \footnote{Code is available on \href{https://anonymous.4open.science/r/TTPO-3651}{Anonymous GitHub}}
\end{abstract}

\input{sections/01-introduction}

\input{sections/02-related-work}

\input{sections/03-method}

\input{sections/04-experiments}

\input{sections/05-conclusion}

\bibliography{custom}

\clearpage

\input{sections/06-appendix}

\end{document}

%% file: sections/01-introduction.tex
\section{Introduction}\label{sec:intro}

Enabling Large Language Models (LLMs)~\citep{gpt4, touvron2023llama2openfoundation} to interact with external environments is critical for enhancing their ability to solve complex real-world problems, such as calling search engines to access real-time information~\citep{NEURIPS2024_e4c61f57} and travel planning~\citep{hao2024large,xie2024travelplanner}. 
As LLMs continue to evolve, integrating external tools is essential not only to address practical user needs but also to advance toward artificial general intelligence~\cite{wang2023mint,liu2023agentbench,tian2024opportunities}.
Current approaches primarily employ Supervised Fine-Tuning (SFT) to improve the tool-use capabilities of LLM~\cite{toolllm,lin2024hammerrobustfunctioncallingondevice,zhang2024adcenhancingfunctioncalling,toolalpaca,toolformer}. 
Although SFT improves tool call quality and facilitates structured outputs, it may still struggle with fine-grained cases where even minor token-level errors can lead to wrong tool calls, such as missing braces. Moreover, they typically rely on synthetic data generated in a forward manner, first creating queries and then generating answers, which may cause many issues, such as unanswerable queries and leakage of tool names or arguments, resulting in low-quality samples that require costly filtering. 
To address the former issue, recent work explores Reinforcement Learning (RL) methods for tool learning, such as TL-Training~\cite{ye2024tltrainingtaskfeaturebasedframeworktraining}, which uses complex reward functions with proximal policy optimization~\cite{schulman2017proximal}, and DPO~\cite{rafailov2023direct}, which leverages trajectory-level preference sampling that focuses on the overall correctness of the full sequence of tool calls. 
While RL-based approaches aim to achieve preference alignment to help models prefer correct tool use, they face key challenges in implementation and stability~\cite{10.1145/3704435}:
(1) Existing methods often \textit{overlook fine-grained preference discrepancies} within individual tool calls, where subtle token-level differences can determine the success or failure of the call.
In highly structured outputs like tool calls, even a single token error can lead to complete failure, highlighting the necessity for more precise preference alignment.
(2) Furthermore, existing preference data sampling methods typically rely on LLM or human evaluations at the trajectory level, rather than assessing each individual tool call. This \textit{coarse-grained assessment} introduces biases due to overlooking fine-grained errors and relying on ambiguous criteria, often resulting in preference data with low discriminative quality and high noise levels, which limits the effectiveness of alignment strategies.

To overcome the above challenges, we propose \fullmodel (\model), a tool-use training paradigm that first constructs token-level preference datasets that align LLMs with fine-grained preferences, and then employs an error-oriented reward mechanism to train the model. 
The proposed \model contains two main components: 
(1)\textit{Preference Oriented Tool-use Dataset Construction}, including \textit{Reversed Dataset Construction} and \textit{Token-level Preference Sampling}, (2)\textit{Error-oriented Scoring Mechanism}. 
In the first component, we first propose a reversed data construction approach, which introduces a novel paradigm for creating multi-turn tool-use datasets to address the latter issue of SFT methods. 
Unlike conventional methods~\citep{toolllm,liu2024toolacewinningpointsllm} that start with queries, our approach reverses the process: we first leverage LLMs to rehearse a sequence of tool calls and a final answer within a predefined tool-using scenario. 
The query is then constructed based on the generated answer. 
This reversed strategy avoids complex and inefficient filtering by deriving queries from scenarios and answers, ensuring each query is answerable and preventing data leakage. Moreover, it maintains question difficulty since the model must use multiple tools, with combined-tool tasks considered more challenging.
To capture the fine-grained preference in the tool calls, we propose \textit{Token-level Preference Sampling}. 
Unlike trajectory-level methods~\citep{NEURIPS2024_c0f7ee19} that incorporate complete tool-calling sequences, our approach explicitly models token-level preferences by sampling top-k candidate tokens from the probability distribution during tool-call generation by LLM. 
When training the tool-use LLM, existing models employ LLMs to grade the outputs as the training signal which usually introduces biases caused by coarse-grained evaluation and ambiguous criteria~\cite{nath2025toolcompmultitoolreasoning}.
Thus, we propose the \textit{Error-oriented Scoring Mechanism}, which defines a taxonomy of tool-call errors.
And then we use it to construct a preference alignment dataset and fine-tune the LLM. 
Extensive experiments on three benchmark datasets show that \model notably improves tool selection, parameter filling, and return value parsing capabilities. 
Moreover, the model fine-tuned with \model demonstrates strong generalization and transferability across datasets, enhancing the reliability and applicability of LLMs in real-world applications.

\noindent Our contributions are summarized as follows:

\noindent $\bullet$ We propose \fullmodel (\model), a novel tool-use training paradigm that aligns the LLM with fine-grained token-level preference to reduce the tool-call errors.

\noindent $\bullet$ We introduce the \textit{Preference Oriented Tool-use Dataset Construction}, which employs a reversed data construction method and a token-level preference sampling approach to construct fine-grained preference data.

\noindent $\bullet$ We propose the \textit{Error-oriented Scoring Mechanism}, which captures fine-grained differences between answers, enabling precise alignment of LLM.

\noindent $\bullet$ Experimental results demonstrate that \model significantly improves tool-use capabilities on three diverse benchmark datasets, and shows strong generalization across models and datasets.

%% file: sections/02-related-work.tex
\section{Related work}

\paragraph{Tool Learning.}
Tool learning enhances LLMs by integrating external tools, enabling them to select tools, generate parameters, and parse results to respond to user queries~\citep{toolw,api-bank,huang2023metatool,shi2023towards}. Approaches include tuning-free methods, which use in-context learning or algorithmic design~\cite{yao2023reactsynergizingreasoningacting,Shi2024ChainOT,huang2024affectsstabilitytoollearning,zhu2025dividethenaggregateefficienttoollearning}, and tuning-based methods, which fine-tune on tool-use datasets~\cite{wu-etal-2024-toolplanner,kong-etal-2024-tptu,gao2024confucius}. Tuning-free methods are often limited by the foundation model's capabilities, while tuning-based methods face challenges with noisy data. Our framework addresses this by employing Reversed Dataset Construction and Token-level Preference Sampling to produce high-quality, low-noise datasets, ensuring better alignment with tool-use tasks and addressing fine-grained discrepancies in tool calls. Additionally, our approach introduces an error-oriented scoring mechanism to refine the alignment process and improve model robustness in complex scenarios.

\paragraph{Tool-Use Datasets.}
Tool learning has driven the creation of datasets to improve LLMs' tool-use capabilities~\cite{patil2023gorilla,wang2024executable,gao2024confucius}. ToolBench~\cite{toolllm} leverages LLMs to compile large datasets, while APIGen~\cite{NEURIPS2024_61cce86d} uses an automated pipeline to generate diverse datasets across multiple API categories. ToolACE~\cite{liu2024toolacewinningpointsllm} further advances this by integrating tool synthesis and dialogue generation, enhancing dataset diversity and complexity. However, these datasets often suffer from noise, single-turn limitations, or high resource costs, and few address the growing need for preference-based datasets. Our framework uses Reversed Dataset Construction and Token-level Preference Sampling to construct high-quality preference datasets, aligning token-level tool-use preferences and improving fine-grained alignment for structured outputs, ensuring better generalization across diverse tool-use scenarios.

%% file: sections/03-method.tex
\section{Method} \label{sec: method}

\subsection{Overview}
\begin{figure*}[!t]
        \centering
	\includegraphics[width=1
 \linewidth]{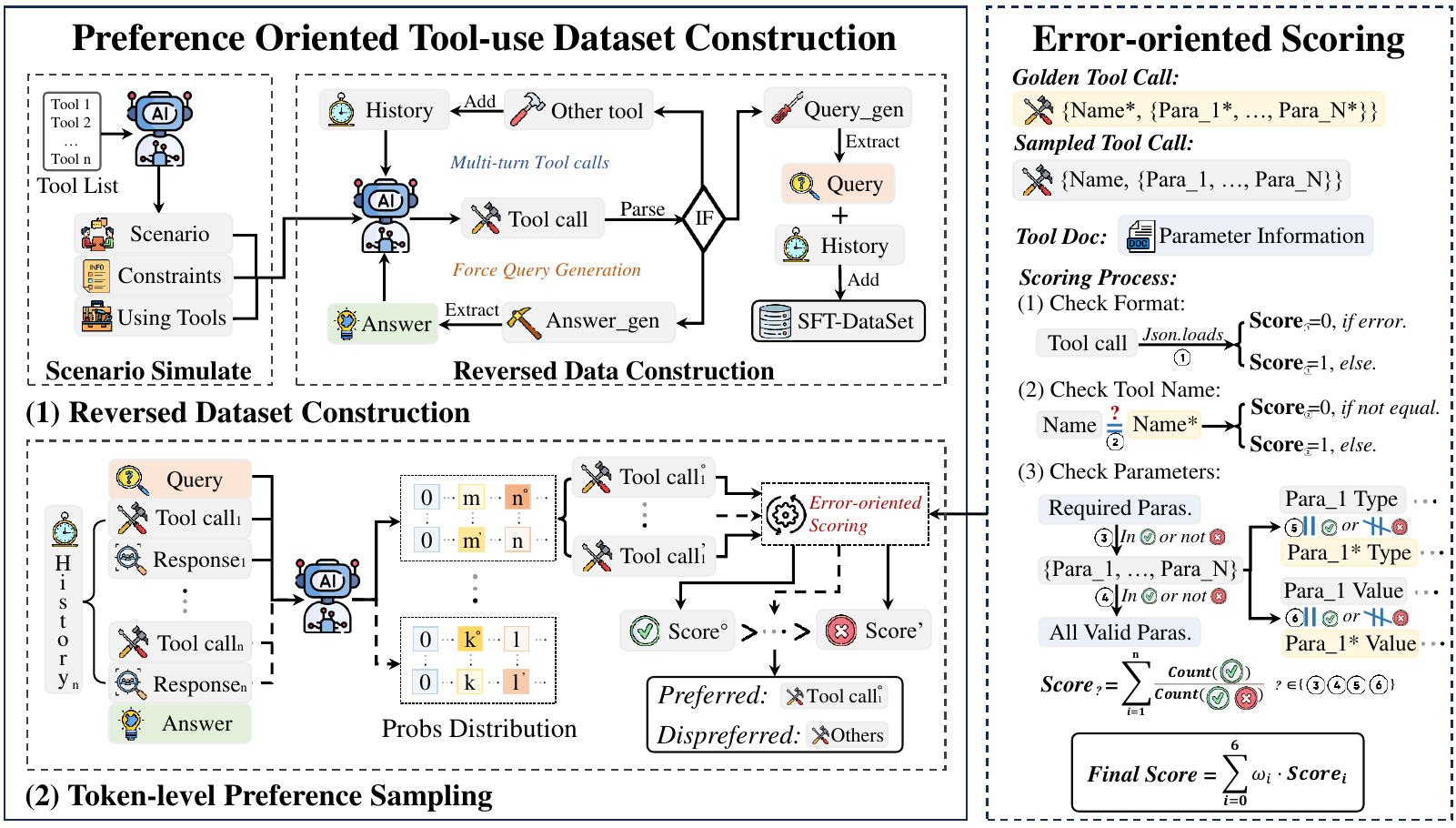}
        \caption{The overall framework of our work, which mainly consists of Preference Oriented Tool-use Dataset Construction and Error-oriented Scoring Mechanism.
        }
 \label{fig:method}
\end{figure*}

In this section, we present the details of the proposed method \model. 
An overview of \model is illustrated in Figure~\ref{fig:method}, which contains two main components:
(1)\textit{Preference Oriented Tool-use Dataset Construction}, a unified framework that includes \textit{Reversed Dataset Construction} for generating reliable and non-leaked raw instruction data, and \textit{Token-level Preference Sampling} for constructing \textit{Preferred} \& \textit{Dispreferred} pairs through fine-grained scoring
(2)\textit{Error-oriented Scoring Mechanism}, a token-level evaluation method designed to capture token-level preferences by fine-grained scoring.
Further details, such as error weights and the example, can be found in Appendix~\ref{sec:app}.

\subsection{Reversed Dataset Construction}

Most existing work trains LLM on synthetic tool-use datasets, and this approach has led to notable progress~\cite{api-bank,toolalpaca,liu2024toolacewinningpointsllm,toolllm,NEURIPS2024_61cce86d}. 
However, in existing tool-use datasets, the generated queries may explicitly reveal information about the tools or parameters involved~\cite{toolllm}. 
In contrast, real-world user queries typically do not explicitly specify the tools to be called or the input parameters. 
This discrepancy creates a gap between the dataset and real-world applications, ultimately affecting the model's performance in practical settings. 
Traditional approaches~\cite{toolllm,liu2024toolacewinningpointsllm} that guide LLMs to first generate a query $Q$ and then solve it, which may result in unsolvable or overly ambiguous queries. While some filtering rules can be applied to remove low-quality data, such filtering consumes significant resources, including API calls, GPU usage, time, and other computational costs, and often fails to achieve the desired effectiveness, as shown in~\cite{NEURIPS2024_61cce86d}.
To address these issues, we propose the \textit{Reversed Dataset Construction} method to construct a tool-use dataset. 

First, we use a candidate tool set $T_{\text{can}}$ as input and then prompt the generator $\mathscr{G}$ to construct three items:
(1) A tool-use scenario description $S$ which is a short sentence to describe this tool-use application scenario.
(2) A toolset $T_{\text{use}} = \{t_1, t_2, \cdots, t_N\}$ with $N$ tools is selected according to the task requirement in the scenario, which should be used in the scenario \textit{S}.
(3) Some constraint $\text{Cons}$ of the scenario \textit{S} to restrict the solution space.
Next, our goal is to generate an answer $A$ based on the tool-use application scenario $S$. 
We simulate the task-solving process by iteratively selecting and calling the tools in $T_{\text{use}}$.
Specifically, in each tool calling step, we predict the tool used in the $i$-th step $t_{\text{call}}^{i}$ based on the answer generation prompt $P_{A}$, the input sequence $S$, the available tools $T_{\text{use}}$, the constraints $Cons$, and the memory of previous tool interactions $M^{i-1}$, where $M^{i-1} = \bigcup_{j}^{i-1} \{ t_{\text{call}}^j, t_{\text{res}}^j \}$. After selecting the tool, we obtain the output $t_{\text{res}}^{i}$ by calling $t_{\text{call}}^{i}$.

After multiple rounds of tool interactions, the generator $\mathscr{G}$ obtains a series of results returned by the tools, and then we generate the answer $A$ according to these inputs.
Finally, we instruct the generator $\mathscr{G}$ to generate a query $Q$.
Since the queries are derived from answers, each query in this dataset is guaranteed to have a valid solution.
Furthermore, the queries, answers, and associated tool results are highly correlated, ensuring that solving the queries necessitates the use of tools. 
This design significantly reduces noise in the dataset, resulting in higher data quality. 

\subsection{Token-level Preference Sampling}
Since the trajectory-level sampling method~\cite{NEURIPS2024_c0f7ee19}, which aligns preferences at a macro level by capturing the overall learning path, usually fails to account for fine-grained distinctions within individual trajectories. 
To tackle this problem, we propose the \textit{Token-level Preference Sampling} strategy for Direct Preference Optimization (DPO). 
For brevity, we denote by $M_{\text{pre}}^{i}$ the set of tool calls and their corresponding return values prior to the $i$-th tool call, i.e., $M_{\text{pre}}^{i} = \{ t_{\text{call}}^{1}, t_{\text{res}}^{1}, \dots, t_{\text{call}}^{i-1}, t_{\text{res}}^{i-1} \}$.
To construct a preference dataset more suitable for training the tool learning model $\mathscr{L}$, we build the dataset by sampling from the outputs of $\mathscr{L}$, which predicts a probability distribution $P_{\text{pred}}$ over possible \textit{tool calls} for the $i$-th step, given the input question $Q$, the available tools $T_{\text{use}}$, and the prior tool usage history $M_{\text{pre}}^{i}$.
During the token-by-token generation process of tool learning model $\mathscr{L}$, the token probability distribution $P_{\text{pred}}$ over the entire vocabulary is computed before each token is generated. 
During sampling, candidate tokens are selected from the top-ranked tokens in $P_{\text{pred}}$.
However, the probability gap between the top-ranked tokens is not always significant, and the probabilities of the top-ranked tokens are very close. 
This close probabilities' distribution creates ambiguity during decoding, as different decoding strategies may randomly select different high-probability tokens. 
Such randomness is particularly problematic for structured and fixed outputs like tool calls, where even a single incorrect token can lead to the failure of the entire tool call.
Therefore, we use the uncertainty in token probabilities as a sampling criterion, perturbing only a small number of tokens at a time to simulate the uncertain sampling behavior of LLMs during the decoding phase:
\begin{gather}
    C_{\text{sam}}^{K} \sim P_{\text{pred}}\mathbb{I}(\text{Dist} < \epsilon ),\\
    \mathrm{where}\quad \text{Dist} = p_{r_1} - p_{r_j},
\end{gather}
where $C_{\text{sam}}^{K}$ denotes $K$-times tool call sampling results in the condition of the distance \textit{Dist} between \textit{rank-j} token's probability $p_{r_j}$ and \textit{rank-1} token's probability $p_{r_1}$ smaller than the predefined hyper-parameter $\epsilon$, the value of $K$ is dynamically determined based on the specific probability. 
Unlike deterministic decoding methods~\cite{shi2024thoroughexaminationdecodingmethods}, which often produce repetitive or suboptimal results, our approach introduces controlled randomness by perturbing a small number of tokens based on their uncertainty. 
Next, we compute the score $\psi_{i}$ for each sampled \textit{tool call} $c_{\text{sam}}^{i} \in C_{\text{sam}}^{K}$ using scoring mechanism $\mathscr{F}$, which can capture fine-grained errors that may occur during tool calls, enabling precise alignment of model preferences. The details of this mechanism will be introduced in \S~\ref{sec:reward}.

\subsection{Error-oriented Scoring Mechanism}\label{sec:reward}

Existing tool learning methods usually employ LLM-based evaluation or human evaluation to assess the quality of generated tool calls, and then use this signal to optimize the model parameters.
In this paper, we design an error-oriented scoring mechanism $\mathscr{F}$ that can capture fine-grained errors that may occur during tool calls.
For tool learning tasks, since tool calls are structured representations, we propose a taxonomy for the tool-call errors.
For a tool call result $ t_{\text{call}} $, the scoring function $ \delta $ is designed to identify whether the call contains errors and to classify these errors into specific error types:
\begin{equation}
 \delta^{e_i}(t_{\text{call}}) = 
\begin{cases}
0, & \text{if} \ e_i\ \text{detected.} \\
1, & \text{if} \ e_i\ \text{not detected.}\\
\end{cases}   
\end{equation}
where $e_i$ denotes a specific error type (\eg format errors and tool name errors).
\input{tables/toolbench}
However, since different tools may have varying numbers of parameters, simply matching the predicted parameters with the ground-truth parameters could result in coarse-grained outcomes.
Therefore, we perform a detailed validation on each parameter output by the model, including type errors and value errors. 
In the evaluation process, each parameter is assigned a score, and the final scores for parameter type errors and parameter value errors are obtained by taking the weighted average of all parameter scores:
\begin{equation}
\delta^{e_i}(t_{\text{call}}) = \frac{1}{X}\sum_j^X{\gamma(v_j)},
\end{equation}
where $\gamma(v_j)$ denotes a similar function to score each parameter $v$ of the $X$ parameters generated by tool learning model $\mathscr{L}$, which can be represented as:
\begin{equation}
\gamma(v_j) = 
\begin{cases}
0, & \text{if} \ v_j\ \text{not correct.} \\
1, & \text{if} \ v_j\ \text{correct.}\\
\end{cases}.
\end{equation}
After the scores for all error types are computed, we obtain the final score for the tool call by weighted sum the scores of all types of errors detecting: 
\begin{equation}
\mathscr{F}(t_{\text{call}}) = \sum_i^H{\omega_i \cdot \delta^{e_i}(t_{\text{call}})},
\end{equation}
where $\omega_i$ denotes the hyper-parameter weight of the type of error $e_i$, $\delta^{e_i}(t_{\text{call}})$ denotes the score of each type of error and $H$ denotes the total number of error types. 
This scoring mechanism can be utilized to generate a preference-aligned dataset, which is subsequently employed for training tool learning models using the DPO method.

%% file: tables/toolbench.tex
\begin{table*}[htbp]
\centering
\begin{adjustbox}{width=2.1\columnwidth,center}
\begin{tabular}{@{} p{3.7cm} >{\centering\arraybackslash}p{1.9cm} >{\centering\arraybackslash}p{1.9cm} >{\centering\arraybackslash}p{1.9cm} >{\centering\arraybackslash}p{1.9cm} >{\centering\arraybackslash}p{1.9cm} >{\centering\arraybackslash}p{1.9cm}@{}}
    \toprule
    \textbf{Models} & \textbf{Vanilla} & \textbf{QS} & \textbf{QL}  & \textbf{TS} & \textbf{TE} & \textbf{TCE}  \\
    \hline
    \multicolumn{7}{l}{\cellcolor{gray!20}\textit{I1-instruction}}        \\
GPT-4o-mini  & 82.0\%          & 80.0\%          & 83.5\%          & 84.0\%          & 81.5\%          & 81.0\%          \\
Hammer2.0-7b & 60.0\%          & 56.0\%          & 54.5\%          & 58.0\%          & 51.5\%          & 53.0\%          \\
xLAM-7b-r    & 77.5\%          & 78.5\%          & 73.5\%          & 79.5\%          & 75.5\%          & 73.0\%          \\
ToolACE-8B   & 77.0\%          & 75.5\%          & 78.5\%          & 74.0\%          & 72.0\%          & 72.0\%          \\
LLaMa3.1-8B & 74.5\% & 74.5\% & 73.5\% & 72.5\% & 71.5\% & 66.5\% \\
Qwen-2.5-7B & 50.0\% & 52.5\% & 45.0\% & 51.5\% & 38.0\% & 40.5\% \\
TTPA (Qwen)     & \textbf{86.0}\% & \textbf{88.5}\% & \textbf{84.5}\% & \textbf{87.5}\% & \textbf{86.0}\% & \textbf{83.5}\%  \\
    \hline
    \multicolumn{7}{l}{\cellcolor{gray!20}\textit{I1-tool}}               \\
GPT-4o-mini  & \textbf{85.5}\% & 83.5\%          & 80.0\%          & 81.5\%          & 83.0\%          & 82.0\%          \\
Hammer2.0-7b & 62.0\%          & 66.0\%          & 56.0\%          & 68.5\%          & 51.0\%          & 51.0\%          \\
xLAM-7b-r    & 77.5\%          & 77.0\%          & 77.0\%          & 73.5\%          & 71.0\%          & 69.5\%          \\
ToolACE-8B   & 76.0\%          & 77.5\%          & \textbf{86.0}\% & 77.5\%          & 76.0\%          & 76.0\%          \\
LLaMa3.1-8B & 77.0\% & 80.5\% & 74.0\% & 77.5\% & 72.0\% & 70.5\% \\
Qwen-2.5-7B & 54.5\% & 60.0\% & 51.0\% & 57.0\% & 42.0\% & 44.5\% \\

TTPA (Qwen)    & 85.0\%          & \textbf{84.0}\% & 82.0\%          & \textbf{81.5}\% & \textbf{83.0}\% & \textbf{83.5}\% \\
    \bottomrule
\end{tabular}
\end{adjustbox}
\caption{The results of evaluation on various ToolBench subsets. The dataset abbreviations correspond to specific modifications: (1) \textbf{Vanilla} represents the original ToolBench dataset; (2) \textbf{Q}uery \textbf{S}horten (QS) denotes the version with condensed queries for increased information density; (3) \textbf{Q}uery \textbf{L}engthen (QL) indicates extended queries with additional information, resulting in sparser key information distribution; (4) \textbf{T}ools \textbf{S}huffle (TS) refers to the variant with randomized tool candidate ordering; (5) \textbf{T}ools \textbf{E}xpand (Intra-category) (TE) represents the expanded toolset within the same category; and (6) \textbf{T}ools \textbf{E}xpand (\textbf{C}ross-category) (TCE) indicates the expanded toolset across different categories. 
We highlight the best performance in \textbf{bold}.
}
\label{tab:toolbench}
\end{table*}

%% file: sections/04-experiments.tex
\section{Experimental Setup}

\subsection{Implementation Details}
To evaluate the effectiveness of \model, we first apply \textit{Reversed Data Construction} and \textit{Token-level Preference Sampling} to generate 3,895 instruction instances and 8,550 preference pairs using 114 specialized APIs.
In this process, we employ state-of-the-art language models, GPT-4o-mini and GPT-4o~\cite{gpt4}, as generators $\mathscr{G}$ to ensure high-quality and valid data.
Subsequently, we fine-tune Qwen2.5-7B-Instruct~\cite{qwen2025qwen25technicalreport} as the tool-use model $\mathscr{L}$ on the constructed dataset to optimize its performance.
\subsection{Baseline}

We conduct a comprehensive comparison between \model and several state-of-the-art baselines in tool use, including: 
(1) \texttt{GPT-4o-mini}, by OpenAI, known for its strong tool-use performance;
(2) \texttt{Hammer2.0-7b}~\cite{lin2024hammerrobustfunctioncallingondevice}, a state-of-the-art tool learning model, demonstrates exceptional function calling capabilities, particularly excelling in robustness.
(3) \texttt{ToolACE-8B}~\cite{liu2024toolacewinningpointsllm}, an advanced tool learning model, trained on coherent dialogue-based tool use datasets for robust multi-turn conversational tool utilization. 
(4) \texttt{xLAM-7b-r}~\cite{NEURIPS2024_61cce86d}, an advanced LLM optimized for decision-making and tool-use from 60k single-turn samples. 
In addition to these models, we also include \texttt{LLaMA-3.1-8B} and \texttt{Qwen-2.5-7B} as baselines in the comparative experiments.
In these baselines, Hammer2.0-7B is fine-tuned from Qwen-2.5-7B, while ToolACE-8B and xLAM-7B-R are based on LLaMA-3.1-8B. Our experiments include models fine-tuned from both the same and different base models, enabling a broader evaluation of TTPA's effectiveness.

\input{tables/datasets}
\input{tables/bfcl}

\subsection{Dataset \& Metric}

We evaluate the tool learning model fine-tuned with \model on two commonly-used benchmarks and our proposed testset. 
The statistics of these datasets are shown in Table~\ref{tab:dataset}.
We first use the subset of widely-used ToolBench~\cite{toolllm} benchmark, including \textit{I1-instruction} and \textit{I1-tool}.
For evaluation, we employ the \textit{Pass Rate} metric, which serves as an intuitive measure of tool learning LLMs' capability in accurately selecting appropriate tools and generating corresponding parameters by the model within a constrained number of inference steps. 
Moreover, we employ the Berkeley Function-Calling Benchmark (BFCL)~\cite{NEURIPS2024_e4c61f57}, which covers complex scenarios such as multiple tool use.
In the evaluation framework, BFCL primarily assesses LLMs based on Abstract Syntax Tree Evaluation. 
This evaluation measures the syntactic correctness of generated tool calls by verifying their alignment with predefined tool documentation in terms of structure and parameters. 
We also employ our testset where we randomly split 10\% of the generated data for testing.
In the testing process, we employ the error-oriented scoring mechanism as the evaluation metric, enabling a fine-grained assessment of tool calls.

\input{tables/ours}

\section{Experimental Result}

\subsection{Overall Performance}

To assess the effectiveness of our proposed \model, we conducted a comprehensive comparison of our model with several strong baseline models across three diverse datasets. 
The results are shown in Table~\ref{tab:toolbench}, Table~\ref{tab:bfcl}, and Table~\ref{tab:ours} for Toolbench, BFCL, and Our testset, respectively. 
\paragraph{ToolBench} 
The findings in ToolBench validate the effectiveness of training on tool-use datasets, revealing that models with merely 7-8 billion parameters can achieve comparable or even superior performance to state-of-the-art GPT-4o-mini in some subsets. 
This highlights the critical role of domain-specific fine-tuning in enhancing the tool-use capability of LLMs. 
Our  \model outperforms the baselines in most scenarios, demonstrating the generalizability of our approach. 
However, an exception is observed in the QL sub-dataset under the I1-tool dataset, where ToolACE-8B achieves better performance. 
This discrepancy can likely be attributed to the fact that ToolACE incorporates extensive dialogue information during its training process, enabling it to handle long queries more effectively. 
Moreover, due to the long-context training data derived from a long candidate tool list, models are required to select the correct tool in more complex scenarios. 
Consequently, our model exhibits higher robustness across five out of six sub-datasets. 
In contrast to other models, where performance fluctuations exceed 5\% even 10\%, our model maintains a pass rate variation of less than 2\%. 
The exception observed in the TCE sub-dataset, where performance declines, is likely due to the crossed expansion of the candidate tool list, which indicates that the model must first identify the appropriate sub-toolsets category before selecting the correct tool within that subset. 
Due to the lack of sufficient training data for this specific challenge, most models perform worse on this dataset compared to their performance on the vanilla dataset. 
Nevertheless, our model still surpasses the baselines, achieving the best performance. 
\paragraph{BFCL} 
The results on BFCL demonstrate that the SOTA baseline models have achieved remarkable performance, particularly on the multiple and simple subsets, where they attain accuracy rates exceeding 90\%. 
Notably, our fine-tuned model demonstrates comparable performance to these existing approaches, reaching SOTA performance levels. 
However, we identify a potential limitation in the BFCL evaluation system: its design may introduce bias during assessment since the number of solutions included for a specific case is fewer than the actual possible solutions. This limitation could lead to two main issues: (1) correct tool calls being misclassified as false, thereby reducing accuracy metrics, and (2) potential favoritism toward models trained on specific datasets that the datas' distribution is similar to the BFCL' data. These factors may partially explain why our model shows slightly inferior performance compared to SOTA models on certain subsets. More detailed case studies can be found in the Appendix~\ref{sec:app:case}.
\paragraph{Our Testset} 
Moreover, on our custom test set, our fine-tuned model outperforms existing advanced tool learning models across three critical aspects that show the capability of tool-use: tool name selection, parameters choosing, and parameters' value filling. 
Specifically, our model achieves accuracies of 57.8\%, 81.3\%, and 74.2\%, respectively, representing at least an average improvement of 11.8\% compared to the baseline advanced models. 
All results on these test sets show the effectiveness of our proposed \model, which can enhance the LLMs' capability of tool-use. 

\begin{figure}[!t]
        \centering
	\includegraphics[width=1
 \linewidth]{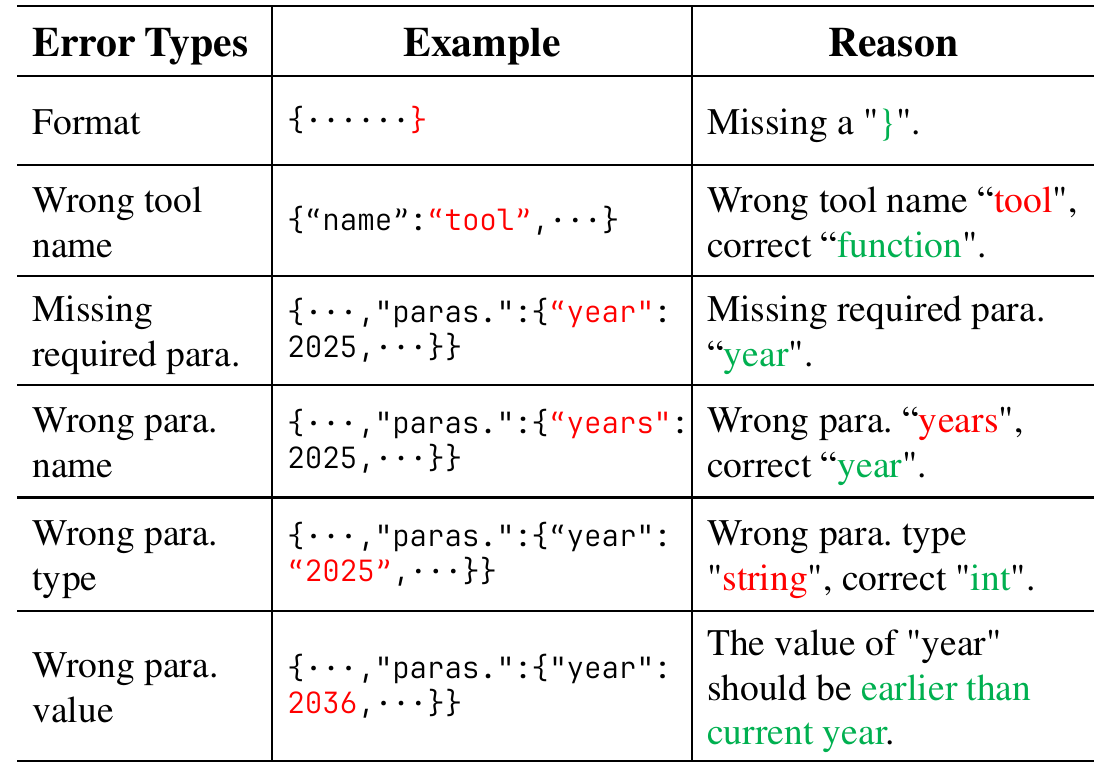}
        \caption{Error types of tool calls. \textit{Example} column presents the examples of different error types. \textit{Reason} column presents the reason why the example failed.
        }
 \label{fig:error}
\end{figure}

\subsection{Error Type Analysis}
In tool-use tasks, LLM errors can be classified into three main categories of six types (Figure~\ref{fig:error})~\cite{dathathri2020plugplaylanguagemodels,ye2024tltrainingtaskfeaturebasedframeworktraining}. Analyzing these errors provides insights for optimizing LLMs' tool-use capabilities.
The first category is format errors, where LLMs must generate machine-parsable tool calls, requiring strict adherence to correct output formats. The second category involves tool selection errors, as LLMs need to choose the most appropriate tool based on task requirements and a thorough understanding of each tool's functionality. The final category concerns parameter errors, which include missing required parameters, invalid parameter types, or values that significantly deviate from the golden references, particularly for parameters involving natural language text.
These errors reflect LLMs' capabilities in three dimensions: (1) instruction following (structured outputs), (2) document comprehension (tool selection), and (3) text generation (parameter filling). This analysis highlights LLMs' limitations and guides targeted improvements in tool-use tasks.

\input{tables/ablation}
\subsection{Ablation Study}
To evaluate the effectiveness of our proposed \model in enhancing the tool-use capabilities of LLMs, we conducted an ablation study comparing the tool-use performance of the base model across various scenarios before and after \model remarkably enhances the tool-use capabilities of LLMs. 
Specifically, we observed substantial improvements across all three benchmark datasets, with performance gains reaching up to 39.7\%. These findings suggest that constructing token-level preference datasets for model fine-tuning enables more granular alignment with correct tool calls while identifying suboptimal or erroneous tool calls, thereby substantially improving tool-use performance.

\begin{figure}[!t]
\centering
\includegraphics[width=0.75\linewidth]{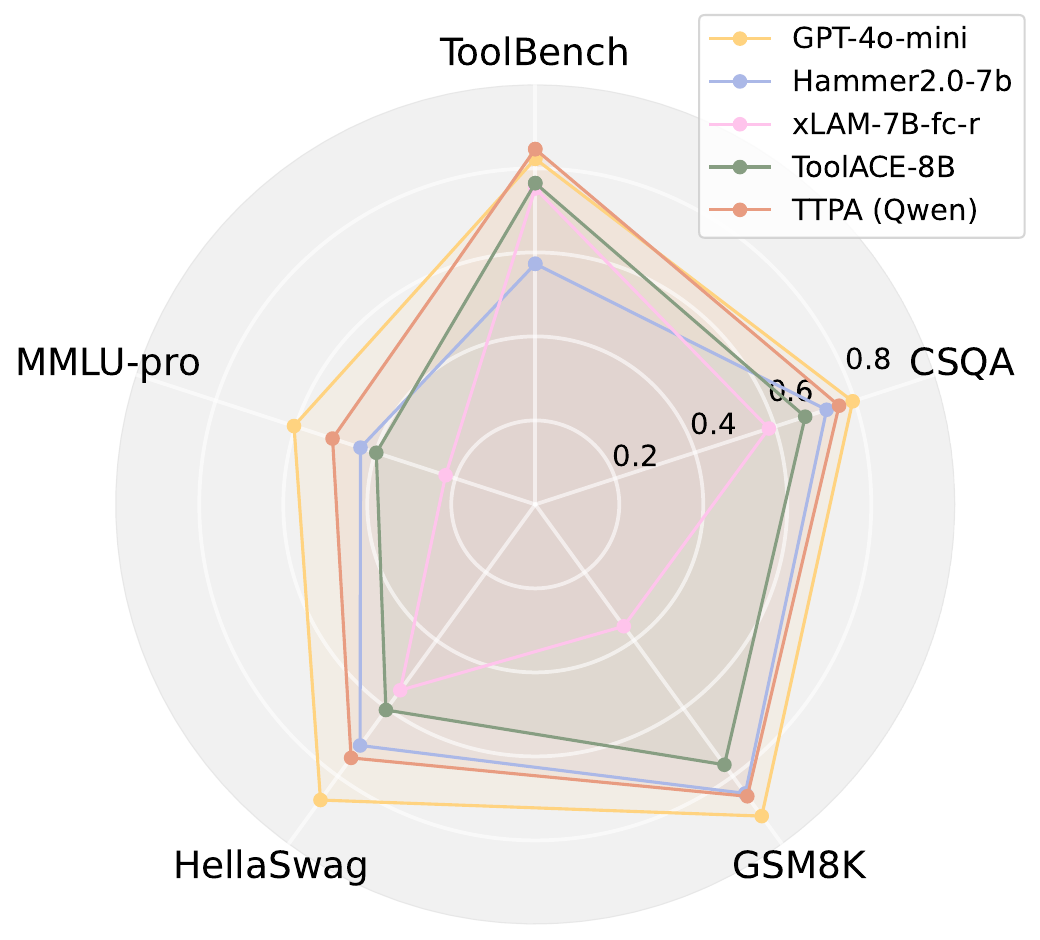}
\caption{The results of evaluation on the general datasets.}
\label{fig:general} 
\end{figure}

\subsection{General Performance}
To comprehensively evaluate the impact of \model on the general capabilities of LLMs, we conduct experiments across multiple benchmarks that assess diverse cognitive abilities: MMLU-pro~\cite{wang2024mmluprorobustchallengingmultitask} fro knowledge mastery, HellaSwag~\cite{zellers-etal-2019-hellaswag} for commonsense reasoning, GSM8K~\cite{cobbe2021trainingverifierssolvemath} for mathematical problem-solving, CommonSenseQA~\cite{talmor-etal-2019-commonsenseqa} for conceptual understanding, and ToolBench for tool-usage. The results, presented in Figure~\ref{fig:general}, demonstrate that the model fine-tuned with \model achieved comparable tool-use capabilities to the state-of-the-art GPT-4o-mini model while maintaining competitive performance across other general benchmarks. Furthermore, our analysis reveals that the model exhibits robust generalization capabilities across different domains, suggesting the effectiveness of the \model fine-tuning approach in both enhancing specialized and maintaining general-purpose performance.

%% file: tables/datasets.tex
\begin{table}[h!]
    \centering
    \begin{adjustbox}{width=0.95\columnwidth,center}
    \begin{tabular}{@{} p{2cm} >{\centering\arraybackslash}p{2.3cm} >{\centering\arraybackslash}p{1.5cm} >{\centering\arraybackslash}p{1.5cm}@{}}
    \toprule
         \textbf{Attributes}&  \textbf{ToolBench} &\textbf{BFCL} &\textbf{Ours}\\ 
         \midrule
         Subsets   & 12   & 5    & 1 \\ 
         Amount    & 2400 & 1929 & 385 \\ 
         APIs      & 1543 &  1100   & 114  \\ 
         Avg. APIs & 5.06    & 1    & 5.56  \\ 
    \bottomrule
     \end{tabular}
     \end{adjustbox}
    \caption{Statistics of the experimental datasets. APIs presents the total number of using APIs in the entire dataset, and Avg. APIs presents the average number of tool-calls per individual case.}
    \label{tab:dataset}
\end{table}

%% file: tables/bfcl.tex
\begin{table*}[htbp]
\centering
\begin{adjustbox}{width=2.05\columnwidth,center}
\begin{tabular}{@{} p{3.6cm} >{\centering\arraybackslash}p{2.3cm} >{\centering\arraybackslash}p{2.3cm} >{\centering\arraybackslash}p{2.3cm} >{\centering\arraybackslash}p{2.3cm} >{\centering\arraybackslash}p{2.3cm} >{\centering\arraybackslash}p{2.3cm}@{}}
    \toprule
    \textbf{Models} & \textbf{Multiple(live)} & \textbf{Simple(live)}  & \textbf{Multiple} & \textbf{Simple} & \textbf{Relevance(live)}\\
    \hline
GPT-4o-mini   & 76.3\%  & 77.1\%   & 90.0\%  & 90.5\% & 77.8\% \\
Hammer2.0-7b   & 75.0\%     & 67.4\%   & 93.5\%   &95.2\% & 83.3\%     \\
xLAM-7b-r     & \textbf{75.4\%}   & 73.6\%   & 95.0\%  & 92.2\% & \textbf{100.0\%}   \\
ToolACE-8B  & 75.2\%    & 78.2\%    & \textbf{95.5\%}   & 95.0\%  & 94.4\% \\
LLaMa-3.1-8B & 65.8\% & 72.8\% & 80.5\% & 91.2\% & 94.4\%   \\
Qwen-2.5-7B  & 72.4\% & 72.6\% & 94.0\%   & 95.3\% & 77.7\% \\

TTPA (Qwen)  & 71.7\%    & \textbf{79.5\%} & 93.0\% & \textbf{95.5\%}   & 94.5\%\\
    \bottomrule
\end{tabular}
\end{adjustbox}
\caption{Accuracy performance on the BFCL subsets. \textit{Multiple} and \textit{Simple} denote that the LLMs are provided multiple tools and one tool, respectively. \textit{live} distinguishes itself from other datasets in the same category. \textbf{Bold} values represent the highest performance for the models evaluated.}
\label{tab:bfcl}
\end{table*}

%% file: tables/ours.tex
\begin{table}[h!]
    \centering
    \begin{adjustbox}{width=0.95\columnwidth,center}
    \begin{tabular}{@{} p{3.1cm} >{\centering\arraybackslash}p{1.4cm} >{\centering\arraybackslash}p{1.4cm} >{\centering\arraybackslash}p{1.4cm}@{}}
    \toprule
         \textbf{Models}&  \textbf{Name} &\textbf{Para.} &\textbf{Content}\\ 
         \midrule
         GPT-4o-mini       & 43.0\% & 70.3\% & 64.6\% \\ 
         Hammer2.0-7b      & 33.9\% & 67.3\% & 59.7\% \\ 
         xLAM-7b-r         & 39.6\% & 71.1\% & 63.1\% \\ 
         ToolACE-8B        & 31.7\% & 62.7\% & 51.1\% \\ 
         LLaMa-3.1-8B        & 32.6\% & 57.1\% & 46.9\% \\ 
         Qwen-2.5-7B        & 29.1\% & 54.4\% & 46.3\% \\
         TTPA (Qwen)         & \textbf{57.8}\% & \textbf{81.3}\% & \textbf{74.2}\% \\ 
    \bottomrule
    \end{tabular}
    \end{adjustbox}
    \caption{Results on our testset. \textit{Name}, \textit{Para.} and \textit{Content} denote the tool calls' accuracy of tool selection, parameters choosing, and parameters content filling, respectively. \textbf{Bold} values represent the highest performance for the models evaluated.}
    \label{tab:ours}
\end{table}

%% file: tables/ablation.tex
\begin{table}[h!]
    \centering
    \begin{adjustbox}{width=0.95\columnwidth,center}
    \begin{tabular}{@{} p{2.9cm} >{\centering\arraybackslash}p{2.4cm} >{\centering\arraybackslash}p{2.4cm}@{}}
    \toprule
\textbf{Dataset}               & \textbf{Base Model} & \textbf{TTPA Model} \\
         \midrule
    \multicolumn{3}{l}{\cellcolor{gray!20}\textit{ToolBench}}        \\
\hspace{1em} -I1-inst.(avg.)                      & 46.3\%                   & \textbf{86.0}\%             \\
\hspace{1em} -I1-tool(avg.)                       & 51.5\%                   & \textbf{83.2}\%             \\
    \multicolumn{3}{l}{\cellcolor{gray!20}\textit{BFCL}}        \\
\hspace{1em} -Multiple(avg.)                & \textbf{83.3}\%              & 82.4\%             \\
\hspace{1em} -Simple(avg.)                  & 84.2\%              & \textbf{87.5}\%             \\
\hspace{1em} -Relevance                     & 77.8\%              & \textbf{94.5}\%             \\
    \multicolumn{3}{l}{\cellcolor{gray!20}\textit{Ours}}        \\
\hspace{1em} -Testset(avg.)                 & 43.3\%    & \textbf{71.1}\%  \\ 
    \bottomrule
    \end{tabular}
    \end{adjustbox}
    \caption{Ablation study. We employ Qwen2.5-7B-Instruct as base model, finetuning with \model. \textit{avg.} presents the average accuracy across all subsets of the corresponding category or different evaluation aspects.}
    \label{tab:ablation}
\end{table}

%% file: sections/05-conclusion.tex
\section{Conclusion}
In this paper, we present \fullmodel (\model), an automated method for constructing high-quality tool-use preference datasets to enhance the tool-use capability of large language models. The \model employs \textit{Preference Oriented Tool-use Dataset Construction}, which incorporates two key components: (1) \textit{Reversed Data Construction} for generating diverse tool-use dataset, and (2) \textit{Token-level Preference Sampling} for capturing token-level preference, to construct a rich and fine-grained tool-use preference dataset that better aligns with real-world usage scenarios. Additionally, we develop an \textit{Error-oriented Scoring Mechanism} that enables precise alignment of LLMs with fine-grained user preferences during tool-usage.
Experiment results demonstrate that the tool learning model fine-tuned with \model can achieve state-of-the-art performance, thereby advancing the field of tool usage in Large Language Models.
\section*{Limitations} 
The main limitation is that conducting fine-grained token-level preference sampling may lead to an increase in computational complexity, requiring higher computational resources and extending the overall training time. In future work, we plan to integrate efficient inference methods with our approach to enhance sampling efficiency. Additionally, our training data is based on a predefined static set of tools, whereas in practical applications, the external environment is dynamically changing. The model's adaptability in dynamic environments still requires further research and validation. We aim to construct a dynamic tool library and extend our method to this dynamic setting, further improving the model's tool-use capabilities in dynamic environments.

\section*{Ethical Considerations}
The research conducted in this paper centers on investigating the effectiveness of fine-grained aligning LLMs for tool-usage. Our work systematically benchmarks LLMs under various real-world scenarios and evaluates their performance. 

In the process of conducting this research, we have adhered to ethical standards to ensure the integrity and validity of our work. All the tasks as well as tools used in our experiment were obtained from existing benchmarks and public open resources, thus ensuring a high level of transparency and reproducibility in our experimental procedure.

To minimize potential bias and promote fairness, we use the prompts following existing works, which are publicly accessible and freely available. We have made every effort to ensure that our research does not harm individuals or groups, nor does it involve any form of deception or potential misuse of information. 

%% file: sections/06-appendix.tex
\appendix
\section{Appendix}\label{sec:app}
\subsection{Case Study}\label{sec:app:case}
\subsubsection{BFCL}
\begin{figure*}[!t]
        \centering
	\includegraphics[width=1
 \linewidth]{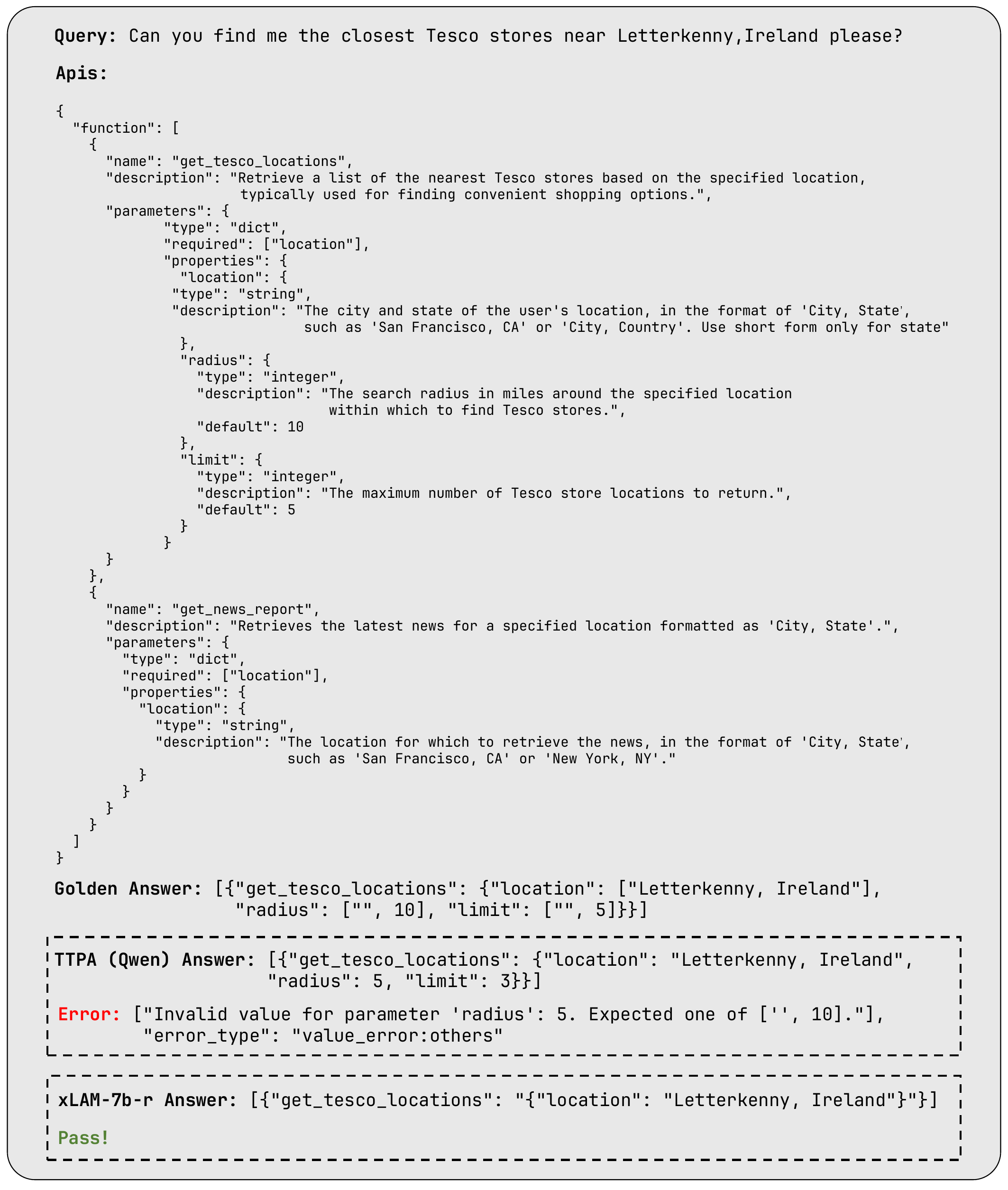}
        \caption{The case study of BFCL. TTPA (Qwen) passes the question but is evaluated as false. 
        }
 \label{fig:case:BFCL}
\end{figure*}

Figure~\ref{fig:case:BFCL} shows one case in the evaluation process of Multiple (live) subset of BFCL datasets, which TTPA (Qwen) failed while xLAM-7b-r success due to the limitation of the evaluate system of BFCL. As shown in Figure~\ref{fig:case:BFCL}, the correct function \textit{get\_tesco\_locations} has three acceptable parameters, where the parameters \textit{radius} and \textit{limit} are optional and not specified. But the golden answer just contains limited valid answers, such that TTPA (Qwen)'s output is evaluated as failure although it generates the correct API name and required parameters (including the parameter's name, type, and value).  

\subsection{Training Details}
The hyper-parameters of the training process are illustrated in Table~\ref{tab:train}. 
\input{tables/train}

\subsection{Error-weights}
The error weight hyperparameters in the Error-oriented Scoring Mechanism are critical since they directly impact the model's performance. In this work, we empirically set the error weights based on preliminary observations. In future work, we plan to conduct a more thorough investigation. 
The error-weights used in the Error-oriented Scoring are detailed in Table~\ref{tab:weights}.

\input{tables/weights}

\subsection{Example of the Entire Process}
The entire process for our proposed model is as follows:

\paragraph{Data Construction:} We first prompt GPT to generate a specific scenario which includes some constraints and using tools. Based on the generated scenario, the generator begins to call the tools multi-turns. Then we generate an answer based on the tool call results and the scenario. Finally, we generate a query corresponding to all the information.

\begin{figure*}[!t]
        \centering
	\includegraphics[width=1
 \linewidth]{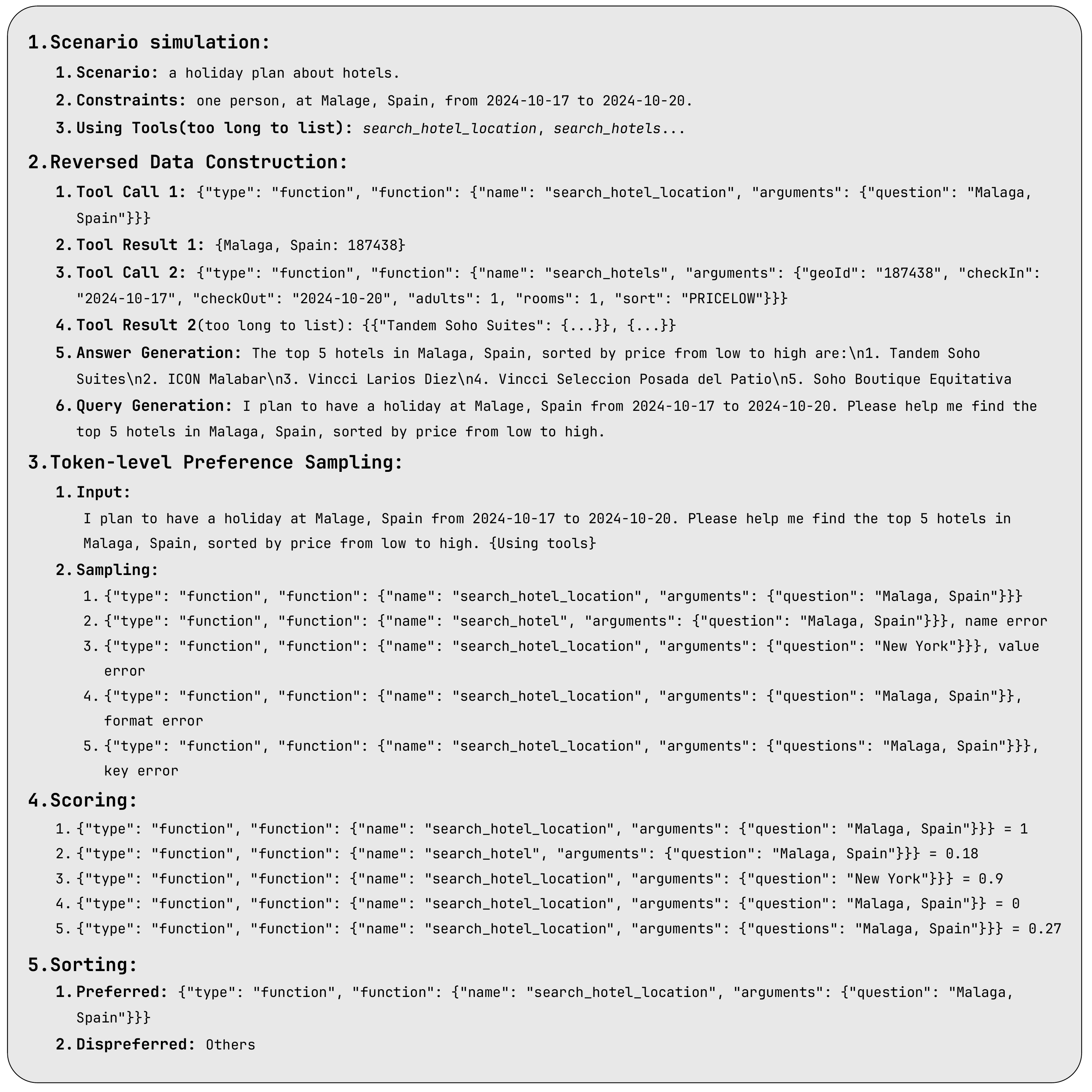}
        \caption{The complete example of entire process.
        }
 \label{fig:example}
\end{figure*}

\paragraph{Tool-Learning:} Then we employ a tool-learning model to solve the generated query. During this process, we sample multiple tool-calling samples from the generated tokens' distribution. By scoring the samples, we can get the preferrence pairs.
A simple example is detailed in Figure~\ref{fig:example}.

\subsection{Details of Proposed Dataset}
The details of the proposed dataset are shown in Table~\ref{tab:pro_data}. 
\input{tables/pro_data}

\subsection{Token-level Analysis}
To capture the changes in sampling ratio during token sampling before and after TTPA training, we design a token-level analysis experiment. Suppose that in each turn, we can sample $x$ tool calls, and solving a problem requires 
$t$ turns. The sampling ratio can be denoted as:

$$
Ratio = \frac{\sum_{i=0}^{t}x}{t}
$$

But in the best situation, the model should generate the right tool call in a high probability and the wrong tool call in a low probability which can not be sampled. So the optimal sampling ratio should be $1$, meaning that the model consistently generates the correct tool call without considering any other alternatives. We can reflect token-level changes by calculating the sampling ratio of each tool call. Specifically, for a given tool call, if there are many possible sampling outcomes, it indicates a higher likelihood of generating incorrect answers. Therefore, this sampling ratio can reflect the token-level changes in model generation after TTPA training, demonstrating the effectiveness of the TTPA training. The specific results are shown in the table~\ref{tab:token-level} below. On our test set, the token distribution generated by the base model is more dispersed compared to that after TTPA training. This is reflected in its higher maximum value, lower mean, and greater variance, indicating instability in model output. This suggests that TTPA training mitigates the likelihood of token-level errors to some extent, as the generated token distribution becomes smoother.

\subsection{Prompt Templates}\label{sec:app:prompt}
The prompts we designed are listed below:

\subsubsection{Reversed Dataset Construction}
Prompt of \textbf{Scenario Simulation:} 
\begin{tcolorbox}[colback=black!1!white,colframe=black!57!white,boxsep=1pt,left=1pt,right=1pt,top=1pt,bottom=1pt,breakable]
\footnotesize
Given the following tools, simulate a scenario where these tools are used in a real-world scenario.\\
You DO NOT need to actually use the tools, just simulate the scenario based on the information provided by the tools.\\
Your goal is to simulate a realistic scenario that involves multiple turns and multiple tools to help another answerer to answer the implicit question asked by a asker.\\

When simulating the scenario, consider the following:\\
1. The scenario should be as realistic as possible and should involve multiple turns (at least two tools).\\
2. The scenario should be related to the tools provided.\\

IMPORTANT: 
The scenario you simulate CAN NOT contain any explicit questions. \\
You SHOULD only state the scenario.\\
The scenario you simulate CAN NOT contain any tool name in the tools above.\\
You SHOULD keep the scenario as realistic as possible. \\

YOUR OUTPUT CONTAINS:\\
scenario: str, the scenario you simulated, it should be a few short words. Also, it should not be a question or instruction. It is just a statement about the scenario.\\

additional\_information: list[str], any information you want to provide about the scenario that may help the answerer to understand the scenario better, at least 4, at most 7. Such as the time, the location, the people involved, etc.\\

tools: list[str], the tools' name  you think are related to the scenario, you should choose the tools from the tools above. And the number of tools should be at least 7, at most 10.\\

There are the tools you can choose:\\
\{tools\}
\end{tcolorbox}

Prompt of \textbf{Answer Generation:} 
\begin{tcolorbox}[colback=black!1!white,colframe=black!57!white,boxsep=1pt,left=1pt,right=1pt,top=1pt,bottom=1pt, breakable]
\footnotesize
You are a data scientist tasked with generating questions to extract specific information from a given dataset.\\
Imagine that there is a asker, you should answer the asker's questions based on the tool calls.\\
But there is no explicit question, you need to answer the implicit question that the asker may have.\\

There are some Steps you can follow:\\
\textbf{Steps:}\\
1. Choose an appropriate tool that you believe can help generate the questions.\\
2. call the selected tool to obtain the tool calls.\\
3. If the tool calls are insufficient to generate the questions, select another tool and repeat the process.\\
4. Once you have gathered enough information, call the Answer\_gen tool to generate an answer based on the tool calls.\\
5. If there are errors, such as the tool returns invalid information or the tool call failed, call the \textbf{Restart} tool to restart.\\

\textbf{Rules:}\\
1. You can choose only one tool at a time.\\
2. The task must involve multiple turns (at least two tools).\\
3. Simulate a realistic scenario in the Additional Information section.\\

\textbf{Additional Information:}\\
\{add\_info\}\\

\textbf{Note:}\\
1. Adapt it to your role and make the task as complex and realistic as possible.\\
2. You should chose the tools related to the scenarios \{scene\} and the information provided.
\end{tcolorbox}

Prompt of \textbf{Query Generation:} 
\begin{tcolorbox}[colback=black!1!white,colframe=black!57!white,boxsep=1pt,left=1pt,right=1pt,top=1pt,bottom=1pt, breakable]
\footnotesize
Imagine that there is a answerer. The answerer answer a question by calling some tools.\\

But there is no explicit question, you need to guess the implicit question that the answerer may answer from the scenario and answer, tool calls given by the answerer.\\

Remember that the implicit question should be closely related to the tool calls and the final answer.\\

But if the answer does not give a clear answer because the tool calls failed, you should guess the implicit question as if the tool calls were successful.\\
Remember that the question should contains the key information that solve the task should be used, such as the date, the location, the people involved, the data to calculate, etc.\\

RULES:\\
1. The question should be designed such that the provided answer is the solution, and the sequence of tool calls represents the steps to derive this answer. \\
2. Ensure the question is intricate and closely related to the tool calls and the final answer. \\
3. Write the question from a first-person perspective, making it sound natural and human-like.\\
4. The question should include the necessary information about the simulation scenario and parameters in a implicit way.
\end{tcolorbox}

The prompts using in the data construction to simulate the user's instructions: 
\begin{tcolorbox}[colback=black!1!white,colframe=black!57!white,boxsep=1pt,left=1pt,right=1pt,top=1pt,bottom=1pt, breakable]
\footnotesize
USER\_PROMPT\_STEP\_1:\\
Please call one tool related to the scenarios: \{choosing\_scenes\}.\\

USER\_PROMPT\_STEP\_2:\\
You can call another tool if you think the tool calls are not enough.\\
Or you can call the Answer\_gen tool to generate the answer based on the tool calls.\\

USER\_PROMPT\_STEP\_3:\\
It's enough. You are allowed to choose at most one another tool expect Answer\_gen tool, then you must call the Answer\_gen tool to generate an answer based on the tool calls.\\

USER\_PROMPT\_STEP\_4:\\
Please generate an answer based on the tool calls.
\end{tcolorbox}

\subsubsection{Token-level Preference Sampling} 
The prompt using in the inference process of the Token-level Preference Sampling: 
\begin{tcolorbox}[colback=black!1!white,colframe=black!57!white,boxsep=1pt,left=1pt,right=1pt,top=1pt,bottom=1pt,breakable]
\footnotesize
You are a tool-use professor, you can use many tools to do the following task that the user ask.\\

At each step, you need to analyze the status now and what to do next, with a tool call to actually execute your step.\\

One step just give one tool call, and you will give ONE step each time I call you. \\

After the call, you will get the call result, and you are now in a new state.\\
Then you will analyze your status now, then decide what to do next...\\
After many steps, you finally perform the task, then you can give your final answer.\\

Remember: \\
1. the state change is irreversible, you can't go back to one of the former state, if you want to restart the task or you want to give the final answer call the Finish tool.\\
2. You can do more then one trys, so if your plan is to continuously try some conditions, you can do one of the conditions per try.\\

Let's Begin!
\end{tcolorbox}

\input{tables/token-level-analysis}

%% file: tables/train.tex
\begin{table*}[h!]
    \centering
    \begin{adjustbox}{width=1.85\columnwidth,center}
    \begin{tabular}{@{} *{7}{>{\centering\arraybackslash}m{2cm}} @{}}
    \toprule
         \textbf{Learning Rate}&  \textbf{Warm-up Ratio} &\textbf{LR Scheduler} &\textbf{Batch Size} &\textbf{Epochs} &\textbf{LoRA rank} &\textbf{LoRA alpha} \\ 
         \midrule
         $10^{-4}$   & 0.1   & cosine    & 32 & 5 & 16 & 32 \\ 
    \bottomrule
     \end{tabular}
     \end{adjustbox}
    \caption{Hyper-parameters in experiments for training.}
    \label{tab:train}
\end{table*}

%% file: tables/weights.tex
\begin{table*}[h!]
    \centering
    \begin{adjustbox}{width=1.35\columnwidth,center}
    \begin{tabular}{@{} *{5}{>{\centering\arraybackslash}m{2cm}} @{}}
    \toprule
         \textbf{Name}&  \textbf{Reqauired Para.} &\textbf{Valid Para.} &\textbf{Para. Type} &\textbf{Para. Value} \\ 
         \midrule
         3   & 3   & 1    & 2 & 2  \\ 
    \bottomrule
     \end{tabular}
     \end{adjustbox}
    \caption{The Error weights used in Error-oriented Scoring.}
    \label{tab:weights}
\end{table*}

%% file: tables/pro_data.tex
\begin{table}[H]
    \centering
    \begin{adjustbox}{width=\columnwidth,center}
    \begin{tabular}{@{} *{5}{>{\centering\arraybackslash}m{1.5cm}} @{}}
    \toprule
         \textbf{Amount}&  \textbf{Domains} &\textbf{APIs} &\textbf{Avg. APIs} &\textbf{Avg. Tokens}\\ 
         \midrule
         3895   & 6   & 114    & 5.56 & 6348\\ 
    \bottomrule
     \end{tabular}
     \end{adjustbox}
    \caption{The details of the proposed dataset. \textit{Avg. APIs} and \textit{Avg. Tokens} denote the average number of API calls and the number of tokens consumed per task, respectively.}
    \label{tab:pro_data}
\end{table}

%% file: tables/token-level-analysis.tex
\begin{table*}[htbp]
\centering
\begin{adjustbox}{width=1.95\columnwidth,center}
\begin{tabular}{@{} p{4.5cm} >{\centering\arraybackslash}p{2cm} >{\centering\arraybackslash}p{2cm} >{\centering\arraybackslash}p{2cm} >{\centering\arraybackslash}p{2cm} >{\centering\arraybackslash}p{2cm}@{}}
    \toprule
    \textbf{Models} & \textbf{Minimum} & \textbf{Maximum} & \textbf{Mean}  & \textbf{Variance} & \textbf{Normalized Variance} \\
    \hline
Qwen2.5-7b-Instruct   & 0     & 1.99   & 0.72   &0.13 & 0.03     \\
TTPA(Qwen2.5-7b-Instruct)      & 0   & 1.62   & 0.80 & 0.08 & 0.01   \\
    \bottomrule
\end{tabular}
\end{adjustbox}
\caption{The token-level analysis experiment to capture the changes of sampling ratio in token sampling before and after TTPA training.}
\label{tab:token-level}
\end{table*}

%% file: TTPA.bbl
\begin{thebibliography}{41}
\providecommand{\natexlab}[1]{#1}

\bibitem[{CHEN et~al.(2024)CHEN, Wang, Wu, Chen, Xu, Luo, Zhang, and Zhang}]{NEURIPS2024_c0f7ee19}
SIJIA CHEN, Yibo Wang, Yi-Feng Wu, Qingguo Chen, Zhao Xu, Weihua Luo, Kaifu Zhang, and Lijun Zhang. 2024.
\newblock Advancing tool-augmented large language models: Integrating insights from errors in inference trees.
\newblock In \emph{Advances in Neural Information Processing Systems}.

\bibitem[{Cobbe et~al.(2021)Cobbe, Kosaraju, Bavarian, Chen, Jun, Kaiser, Plappert, Tworek, Hilton, Nakano, Hesse, and Schulman}]{cobbe2021trainingverifierssolvemath}
Karl Cobbe, Vineet Kosaraju, Mohammad Bavarian, Mark Chen, Heewoo Jun, Lukasz Kaiser, Matthias Plappert, Jerry Tworek, Jacob Hilton, Reiichiro Nakano, Christopher Hesse, and John Schulman. 2021.
\newblock Training verifiers to solve math word problems.
\newblock \emph{arXiv}.

\bibitem[{Dathathri et~al.(2020)Dathathri, Madotto, Lan, Hung, Frank, Molino, Yosinski, and Liu}]{dathathri2020plugplaylanguagemodels}
Sumanth Dathathri, Andrea Madotto, Janice Lan, Jane Hung, Eric Frank, Piero Molino, Jason Yosinski, and Rosanne Liu. 2020.
\newblock Plug and play language models: A simple approach to controlled text generation.
\newblock In \emph{International Conference on Learning Representations: ICLR}.

\bibitem[{Gao et~al.(2024)Gao, Shi, Zhu, Fang, Xin, Ren, Chen, Ma, and Ren}]{gao2024confucius}
Shen Gao, Zhengliang Shi, Minghang Zhu, Bowen Fang, Xin Xin, Pengjie Ren, Zhumin Chen, Jun Ma, and Zhaochun Ren. 2024.
\newblock Confucius: Iterative tool learning from introspection feedback by easy-to-difficult curriculum.
\newblock In \emph{Proceedings of the AAAI Conference on Artificial Intelligence: AAAI}.

\bibitem[{Hao et~al.(2024)Hao, Chen, Zhang, and Fan}]{hao2024large}
Yilun Hao, Yongchao Chen, Yang Zhang, and Chuchu Fan. 2024.
\newblock Large language models can plan your travels rigorously with formal verification tools.
\newblock \emph{arXiv preprint arXiv:2404.11891}.

\bibitem[{Huang et~al.(2024)Huang, Shi, Wen, Chen, Han, Gao, and Shang}]{huang2024affectsstabilitytoollearning}
Chengrui Huang, Zhengliang Shi, Yuntao Wen, Xiuying Chen, Peng Han, Shen Gao, and Shuo Shang. 2024.
\newblock What affects the stability of tool learning? an empirical study on the robustness of tool learning frameworks.
\newblock \emph{arXiv}.

\bibitem[{Huang et~al.(2023)Huang, Shi, Li, Fan, Wu, Zhang, Liu, Zhou, Wan, Gong et~al.}]{huang2023metatool}
Yue Huang, Jiawen Shi, Yuan Li, Chenrui Fan, Siyuan Wu, Qihui Zhang, Yixin Liu, Pan Zhou, Yao Wan, Neil~Zhenqiang Gong, et~al. 2023.
\newblock Metatool benchmark for large language models: Deciding whether to use tools and which to use.
\newblock \emph{arXiv}.

\bibitem[{Kong et~al.(2024)Kong, Ruan, Chen, Zhang, Bao, Shiwei, Qing, Hu, Mao, Li, Zeng, Zhao, and Wang}]{kong-etal-2024-tptu}
Yilun Kong, Jingqing Ruan, YiHong Chen, Bin Zhang, Tianpeng Bao, Shi Shiwei, du~Guo Qing, Xiaoru Hu, Hangyu Mao, Ziyue Li, Xingyu Zeng, Rui Zhao, and Xueqian Wang. 2024.
\newblock {TPTU}-v2: Boosting task planning and tool usage of large language model-based agents in real-world industry systems.
\newblock In \emph{Proceedings of the 2024 Conference on Empirical Methods in Natural Language Processing: Industry Track}.

\bibitem[{Li et~al.(2023)Li, Zhao, Yu, Song, Li, Yu, Li, Huang, and Li}]{api-bank}
Minghao Li, Yingxiu Zhao, Bowen Yu, Feifan Song, Hangyu Li, Haiyang Yu, Zhoujun Li, Fei Huang, and Yongbin Li. 2023.
\newblock {API}-bank: A comprehensive benchmark for tool-augmented {LLM}s.
\newblock In \emph{Association for Computational Linguistics: EMNLP}.

\bibitem[{Lin et~al.(2024)Lin, Wen, Peng, Nie, Liao, Wang, Mo, Zhou, Cheng, Zhao, Wang, and Zhang}]{lin2024hammerrobustfunctioncallingondevice}
Qiqiang Lin, Muning Wen, Qiuying Peng, Guanyu Nie, Junwei Liao, Jun Wang, Xiaoyun Mo, Jiamu Zhou, Cheng Cheng, Yin Zhao, Jun Wang, and Weinan Zhang. 2024.
\newblock Hammer: Robust function-calling for on-device language models via function masking.
\newblock \emph{arXiv}.

\bibitem[{Liu et~al.(2024{\natexlab{a}})Liu, Huang, Zeng, Hao, Yu, Li, Wang, Gan, Liu, Yu, Wang, Wang, Ning, Hou, Wang, Wu, Wang, Liu, Wang, Tang, Tu, Shang, Jiang, Tang, Lian, Liu, and Chen}]{liu2024toolacewinningpointsllm}
Weiwen Liu, Xu~Huang, Xingshan Zeng, Xinlong Hao, Shuai Yu, Dexun Li, Shuai Wang, Weinan Gan, Zhengying Liu, Yuanqing Yu, Zezhong Wang, Yuxian Wang, Wu~Ning, Yutai Hou, Bin Wang, Chuhan Wu, Xinzhi Wang, Yong Liu, Yasheng Wang, Duyu Tang, Dandan Tu, Lifeng Shang, Xin Jiang, Ruiming Tang, Defu Lian, Qun Liu, and Enhong Chen. 2024{\natexlab{a}}.
\newblock Toolace: Winning the points of llm function calling.
\newblock In \emph{International Conference on Learning Representations: ICLR}.

\bibitem[{Liu et~al.(2023)Liu, Yu, Zhang, Xu, Lei, Lai, Gu, Ding, Men, Yang et~al.}]{liu2023agentbench}
Xiao Liu, Hao Yu, Hanchen Zhang, Yifan Xu, Xuanyu Lei, Hanyu Lai, Yu~Gu, Hangliang Ding, Kaiwen Men, Kejuan Yang, et~al. 2023.
\newblock Agentbench: Evaluating llms as agents.
\newblock \emph{arXiv preprint arXiv:2308.03688}.

\bibitem[{Liu et~al.(2024{\natexlab{b}})Liu, Hoang, Zhang, Zhu, Lan, kokane, Tan, Yao, Liu, Feng, R~N, Yang, Savarese, Niebles, Wang, Heinecke, and Xiong}]{NEURIPS2024_61cce86d}
Zuxin Liu, Thai Hoang, Jianguo Zhang, Ming Zhu, Tian Lan, Shirley kokane, Juntao Tan, Weiran Yao, Zhiwei Liu, Yihao Feng, Rithesh R~N, Liangwei Yang, Silvio Savarese, Juan~Carlos Niebles, Huan Wang, Shelby Heinecke, and Caiming Xiong. 2024{\natexlab{b}}.
\newblock Apigen: Automated pipeline for generating verifiable and diverse function-calling datasets.
\newblock In \emph{Advances in Neural Information Processing Systems}.

\bibitem[{Nath et~al.(2025)Nath, Raja, Yoon, and Hendryx}]{nath2025toolcompmultitoolreasoning}
Vaskar Nath, Pranav Raja, Claire Yoon, and Sean Hendryx. 2025.
\newblock Toolcomp: A multi-tool reasoning \& process supervision benchmark.
\newblock \emph{arXiv}.

\bibitem[{OpenAI(2023)}]{gpt4}
OpenAI OpenAI. 2023.
\newblock Gpt-4 technical report.

\bibitem[{Patil et~al.(2023)Patil, Zhang, Wang, and Gonzalez}]{patil2023gorilla}
Shishir~G Patil, Tianjun Zhang, Xin Wang, and Joseph~E Gonzalez. 2023.
\newblock Gorilla: Large language model connected with massive apis.
\newblock \emph{arXiv}.

\bibitem[{Patil et~al.(2024)Patil, Zhang, Wang, and Gonzalez}]{NEURIPS2024_e4c61f57}
Shishir~G Patil, Tianjun Zhang, Xin Wang, and Joseph~E Gonzalez. 2024.
\newblock Gorilla: Large language model connected with massive apis.
\newblock In \emph{Advances in Neural Information Processing Systems}.

\bibitem[{Qin et~al.(2023{\natexlab{a}})Qin, Hu, Lin, Chen, Ding, Cui, Zeng, Huang, Xiao, Han et~al.}]{toolw}
Yujia Qin, Shengding Hu, Yankai Lin, Weize Chen, Ning Ding, Ganqu Cui, Zheni Zeng, Yufei Huang, Chaojun Xiao, Chi Han, et~al. 2023{\natexlab{a}}.
\newblock {Tool learning with foundation models}.
\newblock \emph{arXiv preprint arXiv:2304.08354}.

\bibitem[{Qin et~al.(2024)Qin, Hu, Lin, Chen, Ding, Cui, Zeng, Zhou, Huang, Xiao, Han, Fung, Su, Wang, Qian, Tian, Zhu, Liang, Shen, Xu, Zhang, Ye, Li, Tang, Yi, Zhu, Dai, Yan, Cong, Lu, Zhao, Huang, Yan, Han, Sun, Li, Phang, Yang, Wu, Ji, Li, Liu, and Sun}]{10.1145/3704435}
Yujia Qin, Shengding Hu, Yankai Lin, Weize Chen, Ning Ding, Ganqu Cui, Zheni Zeng, Xuanhe Zhou, Yufei Huang, Chaojun Xiao, Chi Han, Yi~Ren Fung, Yusheng Su, Huadong Wang, Cheng Qian, Runchu Tian, Kunlun Zhu, Shihao Liang, Xingyu Shen, Bokai Xu, Zhen Zhang, Yining Ye, Bowen Li, Ziwei Tang, Jing Yi, Yuzhang Zhu, Zhenning Dai, Lan Yan, Xin Cong, Yaxi Lu, Weilin Zhao, Yuxiang Huang, Junxi Yan, Xu~Han, Xian Sun, Dahai Li, Jason Phang, Cheng Yang, Tongshuang Wu, Heng Ji, Guoliang Li, Zhiyuan Liu, and Maosong Sun. 2024.
\newblock Tool learning with foundation models.
\newblock In \emph{ACM Comput. Surv.}

\bibitem[{Qin et~al.(2023{\natexlab{b}})Qin, Liang, Ye, Zhu, Yan, Lu, Lin, Cong, Tang, Qian, Zhao, Tian, Xie, Zhou, Gerstein, Li, Liu, and Sun}]{toolllm}
Yujia Qin, Shi Liang, Yining Ye, Kunlun Zhu, Lan Yan, Ya-Ting Lu, Yankai Lin, Xin Cong, Xiangru Tang, Bill Qian, Sihan Zhao, Runchu Tian, Ruobing Xie, Jie Zhou, Marc~H. Gerstein, Dahai Li, Zhiyuan Liu, and Maosong Sun. 2023{\natexlab{b}}.
\newblock {ToolLLM: Facilitating Large Language Models to Master 16000+ Real-world APIs}.
\newblock \emph{International Conference on Learning Representations: ICLR}.

\bibitem[{Qwen et~al.(2025)Qwen, :, Yang, Yang, Zhang, Hui, Zheng, Yu, Li, Liu, Huang, Wei, Lin, Yang, Tu, Zhang, Yang, Yang, Zhou, Lin, Dang, Lu, Bao, Yang, Yu, Li, Xue, Zhang, Zhu, Men, Lin, Li, Tang, Xia, Ren, Ren, Fan, Su, Zhang, Wan, Liu, Cui, Zhang, and Qiu}]{qwen2025qwen25technicalreport}
Qwen, :, An~Yang, Baosong Yang, Beichen Zhang, Binyuan Hui, Bo~Zheng, Bowen Yu, Chengyuan Li, Dayiheng Liu, Fei Huang, Haoran Wei, Huan Lin, Jian Yang, Jianhong Tu, Jianwei Zhang, Jianxin Yang, Jiaxi Yang, Jingren Zhou, Junyang Lin, Kai Dang, Keming Lu, Keqin Bao, Kexin Yang, Le~Yu, Mei Li, Mingfeng Xue, Pei Zhang, Qin Zhu, Rui Men, Runji Lin, Tianhao Li, Tianyi Tang, Tingyu Xia, Xingzhang Ren, Xuancheng Ren, Yang Fan, Yang Su, Yichang Zhang, Yu~Wan, Yuqiong Liu, Zeyu Cui, Zhenru Zhang, and Zihan Qiu. 2025.
\newblock Qwen2.5 technical report.
\newblock \emph{arXiv}.

\bibitem[{Rafailov et~al.(2023)Rafailov, Sharma, Mitchell, Manning, Ermon, and Finn}]{rafailov2023direct}
Rafael Rafailov, Archit Sharma, Eric Mitchell, Christopher~D Manning, Stefano Ermon, and Chelsea Finn. 2023.
\newblock Direct preference optimization: Your language model is secretly a reward model.
\newblock \emph{Advances in Neural Information Processing Systems}.

\bibitem[{Schick et~al.(2023)Schick, Dwivedi-Yu, Dess{\`i}, Raileanu, Lomeli, Zettlemoyer, Cancedda, and Scialom}]{toolformer}
Timo Schick, Jane Dwivedi-Yu, Roberto Dess{\`i}, Roberta Raileanu, Maria Lomeli, Luke Zettlemoyer, Nicola Cancedda, and Thomas Scialom. 2023.
\newblock {Toolformer: Language Models Can Teach Themselves to Use Tools}.
\newblock \emph{Neural Information Processing Systems: NeurIPS}.

\bibitem[{Schulman et~al.(2017)Schulman, Wolski, Dhariwal, Radford, and Klimov}]{schulman2017proximal}
John Schulman, Filip Wolski, Prafulla Dhariwal, Alec Radford, and Oleg Klimov. 2017.
\newblock Proximal policy optimization algorithms.
\newblock \emph{arXiv preprint arXiv:1707.06347}.

\bibitem[{Shi et~al.(2024{\natexlab{a}})Shi, Yang, Cai, Zhang, Wang, Yang, and Lam}]{shi2024thoroughexaminationdecodingmethods}
Chufan Shi, Haoran Yang, Deng Cai, Zhisong Zhang, Yifan Wang, Yujiu Yang, and Wai Lam. 2024{\natexlab{a}}.
\newblock A thorough examination of decoding methods in the era of llms.
\newblock \emph{arXiv}.

\bibitem[{Shi et~al.(2024{\natexlab{b}})Shi, Gao, Chen, Feng, Yan, Shi, Yin, Chen, Verberne, and Ren}]{Shi2024ChainOT}
Zhengliang Shi, Shen Gao, Xiuyi Chen, Yue Feng, Lingyong Yan, Haibo Shi, Dawei Yin, Zhumin Chen, Suzan Verberne, and Zhaochun Ren. 2024{\natexlab{b}}.
\newblock Chain of tools: Large language model is an automatic multi-tool learner.
\newblock \emph{ArXiv}.

\bibitem[{Shi et~al.(2023)Shi, Gao, Zhang, Chen, Chen, Ren, and Ren}]{shi2023towards}
Zhengliang Shi, Shen Gao, Zhen Zhang, Xiuying Chen, Zhumin Chen, Pengjie Ren, and Zhaochun Ren. 2023.
\newblock Towards a unified framework for reference retrieval and related work generation.
\newblock In \emph{Association for Computational Linguistics: EMNLP}.

\bibitem[{Talmor et~al.(2019)Talmor, Herzig, Lourie, and Berant}]{talmor-etal-2019-commonsenseqa}
Alon Talmor, Jonathan Herzig, Nicholas Lourie, and Jonathan Berant. 2019.
\newblock {C}ommonsense{QA}: A question answering challenge targeting commonsense knowledge.
\newblock In \emph{Proceedings of the 2019 Conference of the North {A}merican Chapter of the Association for Computational Linguistics: Human Language Technologies, Volume 1 (Long and Short Papers)}.

\bibitem[{Tang et~al.(2023)Tang, Deng, Lin, Han, Liang, and Sun}]{toolalpaca}
Qiaoyu Tang, Ziliang Deng, Hongyu Lin, Xianpei Han, Qiao Liang, and Le~Sun. 2023.
\newblock {Toolalpaca: Generalized tool learning for language models with 3000 simulated cases}.
\newblock \emph{arXiv preprint arXiv:2306.05301}.

\bibitem[{Tian et~al.(2024)Tian, Jin, Yeganova, Lai, Zhu, Chen, Yang, Chen, Kim, Comeau et~al.}]{tian2024opportunities}
Shubo Tian, Qiao Jin, Lana Yeganova, Po-Ting Lai, Qingqing Zhu, Xiuying Chen, Yifan Yang, Qingyu Chen, Won Kim, Donald~C Comeau, et~al. 2024.
\newblock Opportunities and challenges for chatgpt and large language models in biomedicine and health.
\newblock In \emph{Briefings in Bioinformatics}.

\bibitem[{Touvron et~al.(2023)Touvron, Martin, Stone, Albert, Almahairi, Babaei, Bashlykov, Batra, Bhargava, Bhosale, Bikel, Blecher, Ferrer, Chen, Cucurull, Esiobu, Fernandes, Fu, Fu, Fuller, Gao, Goswami, Goyal, Hartshorn, Hosseini, Hou, Inan, Kardas, Kerkez, Khabsa, Kloumann, Korenev, Koura, Lachaux, Lavril, Lee, Liskovich, Lu, Mao, Martinet, Mihaylov, Mishra, Molybog, Nie, Poulton, Reizenstein, Rungta, Saladi, Schelten, Silva, Smith, Subramanian, Tan, Tang, Taylor, Williams, Kuan, Xu, Yan, Zarov, Zhang, Fan, Kambadur, Narang, Rodriguez, Stojnic, Edunov, and Scialom}]{touvron2023llama2openfoundation}
Hugo Touvron, Louis Martin, Kevin Stone, Peter Albert, Amjad Almahairi, Yasmine Babaei, Nikolay Bashlykov, Soumya Batra, Prajjwal Bhargava, Shruti Bhosale, Dan Bikel, Lukas Blecher, Cristian~Canton Ferrer, Moya Chen, Guillem Cucurull, David Esiobu, Jude Fernandes, Jeremy Fu, Wenyin Fu, Brian Fuller, Cynthia Gao, Vedanuj Goswami, Naman Goyal, Anthony Hartshorn, Saghar Hosseini, Rui Hou, Hakan Inan, Marcin Kardas, Viktor Kerkez, Madian Khabsa, Isabel Kloumann, Artem Korenev, Punit~Singh Koura, Marie-Anne Lachaux, Thibaut Lavril, Jenya Lee, Diana Liskovich, Yinghai Lu, Yuning Mao, Xavier Martinet, Todor Mihaylov, Pushkar Mishra, Igor Molybog, Yixin Nie, Andrew Poulton, Jeremy Reizenstein, Rashi Rungta, Kalyan Saladi, Alan Schelten, Ruan Silva, Eric~Michael Smith, Ranjan Subramanian, Xiaoqing~Ellen Tan, Binh Tang, Ross Taylor, Adina Williams, Jian~Xiang Kuan, Puxin Xu, Zheng Yan, Iliyan Zarov, Yuchen Zhang, Angela Fan, Melanie Kambadur, Sharan Narang, Aurelien Rodriguez, Robert Stojnic, Sergey Edunov, and Thomas
  Scialom. 2023.
\newblock Llama 2: Open foundation and fine-tuned chat models.
\newblock \emph{arXiv}.

\bibitem[{Wang et~al.(2024{\natexlab{a}})Wang, Chen, Yuan, Zhang, Li, Peng, and Ji}]{wang2024executable}
Xingyao Wang, Yangyi Chen, Lifan Yuan, Yizhe Zhang, Yunzhu Li, Hao Peng, and Heng Ji. 2024{\natexlab{a}}.
\newblock Executable code actions elicit better llm agents.
\newblock \emph{arXiv preprint arXiv:2402.01030}.

\bibitem[{Wang et~al.(2023)Wang, Wang, Liu, Chen, Yuan, Peng, and Ji}]{wang2023mint}
Xingyao Wang, Zihan Wang, Jiateng Liu, Yangyi Chen, Lifan Yuan, Hao Peng, and Heng Ji. 2023.
\newblock Mint: Evaluating llms in multi-turn interaction with tools and language feedback.
\newblock \emph{International Conference on Learning Representations: ICLR}.

\bibitem[{Wang et~al.(2024{\natexlab{b}})Wang, Ma, Zhang, Ni, Chandra, Guo, Ren, Arulraj, He, Jiang, Li, Ku, Wang, Zhuang, Fan, Yue, and Chen}]{wang2024mmluprorobustchallengingmultitask}
Yubo Wang, Xueguang Ma, Ge~Zhang, Yuansheng Ni, Abhranil Chandra, Shiguang Guo, Weiming Ren, Aaran Arulraj, Xuan He, Ziyan Jiang, Tianle Li, Max Ku, Kai Wang, Alex Zhuang, Rongqi Fan, Xiang Yue, and Wenhu Chen. 2024{\natexlab{b}}.
\newblock Mmlu-pro: A more robust and challenging multi-task language understanding benchmark.
\newblock \emph{arXiv}.

\bibitem[{Wu et~al.(2024)Wu, Liu, Luan, and Wang}]{wu-etal-2024-toolplanner}
Qinzhuo Wu, Wei Liu, Jian Luan, and Bin Wang. 2024.
\newblock {T}ool{P}lanner: A tool augmented {LLM} for multi granularity instructions with path planning and feedback.
\newblock In \emph{Proceedings of the 2024 Conference on Empirical Methods in Natural Language Processing}.

\bibitem[{Xie et~al.(2024)Xie, Zhang, Chen, Zhu, Lou, Tian, Xiao, and Su}]{xie2024travelplanner}
Jian Xie, Kai Zhang, Jiangjie Chen, Tinghui Zhu, Renze Lou, Yuandong Tian, Yanghua Xiao, and Yu~Su. 2024.
\newblock Travelplanner: A benchmark for real-world planning with language agents.
\newblock \emph{arXiv preprint arXiv:2402.01622}.

\bibitem[{Yao et~al.(2023)Yao, Zhao, Yu, Du, Shafran, Narasimhan, and Cao}]{yao2023reactsynergizingreasoningacting}
Shunyu Yao, Jeffrey Zhao, Dian Yu, Nan Du, Izhak Shafran, Karthik Narasimhan, and Yuan Cao. 2023.
\newblock React: Synergizing reasoning and acting in language models.
\newblock In \emph{International Conference on Learning Representations: ICLR}.

\bibitem[{Ye et~al.(2024)Ye, Wu, Li, Yang, Gui, Zhang, Huang, Wang, Shi, Fan, and Du}]{ye2024tltrainingtaskfeaturebasedframeworktraining}
Junjie Ye, Yilong Wu, Sixian Li, Yuming Yang, Tao Gui, Qi~Zhang, Xuanjing Huang, Peng Wang, Zhongchao Shi, Jianping Fan, and Zhengyin Du. 2024.
\newblock Tl-training: A task-feature-based framework for training large language models in tool use.
\newblock \emph{arXiv}.

\bibitem[{Zellers et~al.(2019)Zellers, Holtzman, Bisk, Farhadi, and Choi}]{zellers-etal-2019-hellaswag}
Rowan Zellers, Ari Holtzman, Yonatan Bisk, Ali Farhadi, and Yejin Choi. 2019.
\newblock {H}ella{S}wag: Can a machine really finish your sentence?
\newblock In \emph{Proceedings of the 57th Annual Meeting of the Association for Computational Linguistics}.

\bibitem[{Zhang et~al.(2024)Zhang, Zhang, Zhu, Jia, Jiang, Guo, Li, and Zhou}]{zhang2024adcenhancingfunctioncalling}
Wei Zhang, Yi~Zhang, Li~Zhu, Qianghuai Jia, Feijun Jiang, Hongcheng Guo, Zhoujun Li, and Mengping Zhou. 2024.
\newblock Adc: Enhancing function calling via adversarial datasets and code line-level feedback.

\bibitem[{Zhu et~al.(2025)Zhu, Shi, Shi, Ren, Wang, Yan, and Yin}]{zhu2025dividethenaggregateefficienttoollearning}
Dongsheng Zhu, Weixian Shi, Zhengliang Shi, Zhaochun Ren, Shuaiqiang Wang, Lingyong Yan, and Dawei Yin. 2025.
\newblock Divide-then-aggregate: An efficient tool learning method via parallel tool invocation.
\newblock \emph{arXiv}.

\end{thebibliography}
